\documentclass[runningheads]{llncs}
\usepackage{amsmath}
\usepackage[T1]{fontenc}
\usepackage{subcaption}
\usepackage{graphicx}
\usepackage{verbatim}
\usepackage{booktabs}
\usepackage{subcaption}
\usepackage{makecell}

\usepackage[normalem]{ulem} % Add this line to your preamble
\usepackage[table,xcdraw]{xcolor} % Assuming you're using this for cell coloring
\usepackage{multirow}
\usepackage{adjustbox}
\usepackage[table,xcdraw]{xcolor}
\usepackage{svg}
\usepackage{hyperref}

\usepackage{booktabs}

\begin{document}
\title{Large Language Models Assisting Ontology Evaluation}
\def\thefootnote{*}\footnotetext{Equal contribution.}

\def\thefootnote{*}\footnotetext{Equal contribution.}
\author{Anna Sofia Lippolis$^{1,2,*}$ \orcidID{0000-0002-0266-3452} \and
Mohammad Javad Saeedizade$^{3,*}$ \orcidID{0009-0000-0812-6167} \and
Robin Keskisärkkä\inst{3}\orcidID{0000-0003-0644-4051} \and
Aldo Gangemi\inst{1,2}\orcidID{0000-0001-5568-2684} \and
Eva Blomqvist\inst{3}\orcidID{0000-0003-0036-6662} \and 
Andrea Giovanni Nuzzolese\inst{2}\orcidID{0000-0003-2928-9496}
}
\authorrunning{A. S. Lippolis et al.}

\institute{
  University of Bologna, Italy \\
  \email{annasofia.lippolis2@unibo.it, aldo.gangemi@unibo.it}
  \and
  ISTC-CNR, Italy \\
  \email{andrea.nuzzolese@cnr.it}
  \and
  Linköping University, Sweden \\
  \email{javad.saeedizade@liu.se, robin.keskisarkka@liu.se, eva.blomqvist@liu.se}
}
\maketitle              % typeset the header of the contribution
\begin{abstract}
Ontology evaluation through functional requirements---such as testing via competency question (CQ) verification---is a well-established yet costly, labour-intensive, and error-prone endeavour, even for ontology engineering experts. In this work, we introduce \textbf{OE-Assist}, a novel framework designed to assist ontology evaluation through automated and semi-automated CQ verification. By presenting and leveraging a dataset of 1,393 CQs paired with corresponding ontologies and ontology stories, our contributions present, to our knowledge, the first systematic investigation into large language model (LLM)-assisted ontology evaluation, and include: (i) evaluating the effectiveness of a LLM-based approach for automatically performing CQ verification against a manually created gold standard, and (ii) developing and assessing an LLM-powered framework to assist CQ verification with Protégé, by providing suggestions. We found that \textit{automated} LLM-based evaluation with o1-preview and o3-mini perform at a similar level to the average user's performance. % on LLM-generated CQs, while the evaluation on human-authored ontologies score lower.
%, suggesting that newer models better align with the style of LLM-generated content.
Through a user study on the framework with 19 knowledge engineers from eight international institutions, we also show that LLMs can assist \textit{manual} CQ verification and improve user accuracy, especially when suggestions are correct. Additionally, participants reported a marked decrease in perceived task difficulty. However, we also observed a reduction in human performance when the LLM provided incorrect guidance, showing a critical trade-off between efficiency and accuracy in assisted ontology evaluation. %Given that correct suggestions outnumber incorrect ones, \textbf{OE-Assist} ultimately enhances overall performance compared to traditional workflows.
%Things to mention:

%1- Ontology evaluation using CQ verification is time-consuming and costly

%2- It is not a straightforward task so oncologists make mistakes since different concepts can mean different to different people and finding them is hard

%3- we created a dataset of around 1500 CQs and corresponding ontologists with human annotations.

%4- we created a tool and evaluated it with ontologists.

%5- our findings show

%6- we are the first to work in this field.

\end{abstract}
\renewcommand*{\thefootnote}{\arabic{footnote}}
\setcounter{footnote}{0}

\section{Introduction}
\label{sec:intro}

\begin{comment}
   Things to say:

paragraphs 1-2 marketing the work :)

Our contribution + findings?

The structure of paper

Outline
1. CQs as a basis for ontology evaluation
2. challenges in ontology evaluation using CQs
- time and resource intensive
- complexity and ambiguity (difficulty of standardizing and interpreting ontological concepts)
3. KE automation by ontology engineers
- recent interest in applyign LLMs to ke
- variability in evaluation measures
- to date, none have focused on LLM-assisted evaluation for CQ verification.
4. Our work
- to fill this gap, our work is a novel framework designed to automate the functional evaluation of ontologies through CQ verification
- building on previous work we build a dataset for ontology evaluation
- ... 
\end{comment}

Competency questions~\cite{gruninger1995role} (CQs), explicit questions an ontology should be able to answer, have long been considered essential functional requirements in the ontology engineering lifecycle, especially in iterative ontology development frameworks such as eXtreme Design~\cite{presutti2009extreme}. Functional ontology testing through CQ verification, e.g., by formulating the question as a SPARQL query as suggested in ~\cite{blomqvist2012ontology}, has been applied in knowledge engineering for more than a decade (e.g. \cite{carriero2019arco,lippolis2025ontology,litovkin2020suitability,saeedizade2024navigating,ontochat2024,zhang2025trustworthy}). In this way, it is possible to capture actual inferential capability and the ontology's alignment with its design goals. Despite its many advantages, the evaluation of ontologies through CQ verification poses several challenges. First, the process is both time- and resource-intensive, as it requires different kinds of expertise and is entirely manual. Second, it is highly dependent on the complexity, domain-specificity, and ambiguity of the terms used, as standardizing and interpreting ontological concepts is often difficult. Recent interest in applying large language models (LLMs) to knowledge engineering underscores their potential role in automating different steps of the ontology engineering lifecycle, such as formulating or retrofitting CQs~\cite{alharbi2024experiment,alharbi2024review,antia2023automating,ontochat2024} and generating ontologies from CQs and user stories~\cite{lippolisontogenia,lippolis2025ontology,saeedizade2024navigating}. For what concerns ontology generation, the lack of consistency in evaluation measures makes it difficult to establish a standardised framework for assessing ontology generation. Furthermore, no study has yet explored LLMs' potential for  functional evaluation of ontologies through direct CQ verification, especially in a semi-automated setting that includes human-in-the-loop.

To fill this gap, our work presents \textbf{OE-Assist}, a novel framework designed to automate the functional evaluation of ontologies through CQ verification. The ultimate goal of our approach is to provide the end users with LLM-based suggestions to verify the modelling of ontologies according to requirements. The CQ verification method is particularly well-suited for LLM assistance as: (1) CQs are expressed in natural language, aligning with LLMs' core capabilities; (2) the verification process involves reasoning about the semantic alignment between requirements and implementation; and (3) the translation between CQs and SPARQL queries represents precisely the kind of symbolic-natural language boundary that LLMs are designed to navigate.
Our method involves using an LLM to generate, for a given ontology, both verification labels (\textsc{Yes}/\textsc{No}) and a SPARQL query that motivates the label, confirming whether a CQ has been correctly modelled. %This approach thus integrates the LLM's suggestions with the ontology to be evaluated.

Expanding on a dataset of LLM-generated ontologies from previous work ~\cite{lippolis2025ontology}, we developed a dataset of 1,393 CQs, stories, and corresponding ontologies for ontology evaluation, whose assessment was cross-checked by two ontology engineers. This resource was used to investigate the performance of an LLM-based evaluation approach, against annotations in the dataset (the gold standard). Furthermore, based on this larger dataset, we derived a balanced dataset to test the efficacy of an interface prototype. It provides LLM-based suggestions to 19 ontology engineers of varying expertise and affiliation during CQ verification of an ontology with Protégé\footnote{\url{https://protege.stanford.edu/}}, with and without suggestions. This paper aims to answer the following research questions:

%This prototype, with and without suggestions, was evaluated by 20 knowledge engineers of varying expertise—from student-level to experts—from eight different international institutions.

% To what extent can LLMs evaluate ontologies using CQ verification?
% To what extent can LLMs can assist ontology engineers to evaluate ontologies using CQ verification? 

\begin{itemize}
    \item \textbf{RQ1.} To what extent can LLMs evaluate ontologies using CQ verification?

    \item \textbf{RQ2.} To what extent can LLMs assist ontology engineers in evaluating ontologies through CQ verification, and what are the benefits and drawbacks of a hybrid approach combining LLM suggestions with expert validation compared to traditional human-only methods?
    %To what extent can LLMs can assist ontology engineers to evaluate ontologies using CQ verification? (What are the benefits and drawbacks of a hybrid evaluation approach that combines LLM suggestions with human expert confirmation, compared to traditional human-only methods?)
    % \begin{itemize}
    %     \item \textbf{RQ3.1} Is there a drop in accuracy from using LLM suggestions compared to traditional human-only methods?
    %     \item \textbf{RQ3.2} Does perceived difficulty change using LLM suggestions compared to traditional human-only methods?
    %     \item \textbf{RQ3.3} Is there a difference in time spent from using LLM suggestions compared to traditional human-only methods?
    %     \item \textbf{RQ3.4} Is there a difference between highly experienced vs low-experienced groups? 
    % \end{itemize}
\end{itemize}
\vspace{-0.5em}

The remainder of the paper is organized as follows: 
Section~\ref{sec:relatedWork} reviews relevant literature; Section~\ref{sec:dataset} describes the OntoEval dataset; Section \ref{sec:methodology} shows our research methodology and describes our experimental setup; Section~\ref{sec:results} presents our findings; Section~\ref{sec:discussion} discusses the implications of the results. Finally, Section~\ref{sec:conclusion} summarizes the main findings.

\section{Related Work}
\label{sec:relatedWork}
The first mention of CQs for ontology development cycles can be traced back to 1996, where Uschold and Gruninger described CQs as ontology requirements bridging the gap between informal questions and formal queries~\cite{uschold1996ontologies}. %In this view, CQs are natural language queries an ontology must be able to answer to evaluate the expressiveness of the ontology and constrain the ontology design search space~\cite{noy1997state}.
Later, Vrandečić and Gangemi~\cite{vrandevcic2006unit} proposed the treatment of ontology requirements as unit tests, according to which each CQ should be turned into a query whose results on the ontology can be checked against expected outcomes. This work introduced CQ-driven validation as a distinct evaluation dimension (termed \textit{functional evaluation}), related to the intended use of an ontology and separate from structural analysis~\cite{gangemi2006modelling}. %In this way, ontology validation can be considered a diagnostic task, with fitness-to-competency questions constraining the intended use for the ontology.
%Such competency-driven validation could then be treated as a distinct evaluation dimension (termed \textit{functional evaluation}), related to the intended use of an ontology and its components, i.e. their function, and separate from structural analysis \cite{gangemi2006modelling}. The authors propose functional measures to include competence adequacy (e.g. inter-subjective agreement, task adequacy, task specifity, and topic specificity); NLP adequacy (e.g. compliance with lexical distinctions), and functional modularity (e.g. ontology stratification, or granularity). As such, ontology validation is outlined as a diagnostic task, including the requirement to be able to answer a certain competency question, and the fitness-to-competency questions constrain the setting of the intended use for the ontology.
%According to the classification in Gangemi et al. \cite{gangemi2006modelling}, the current work on ontology evaluation mostly concerns the structural dimension of ontology (e.g. syntax and formal semantics, logical-adequacy properties), but not as much the functional dimension. 
The importance of CQs was further cemented in main agile ontology engineering methodologies like \textsc{Methontology}~\cite{fernandez1997methontology} and eXtreme Design~\cite{blomqvist2010experimenting}, which included CQs as part of the requirement specification and verification phases. Building on this foundation, Blomqvist et al.~\cite{blomqvist2012ontology} proposed a methodology for ontology testing that uses CQs as requirements. They showed that reformulating CQs into SPARQL queries effectively identifies mistakes. In fact, CQ verification has been used for over a decade in several projects~\cite{carriero2019arco,lippolis2025ontology,litovkin2020suitability,saeedizade2024navigating,ontochat2024,zhang2025trustworthy}.
%, and researchers have analyzed large sets of real competency questions to understand common patterns and facilitate their formalization \cite{keet2024discerning}. 
At the same time, manually writing these queries is perceived as tedious work, pointing directly to the automation opportunity our work addresses.

%Despite widespread adoption, CQ verification challenges identified in these early works persist today.  However, the manual effort required for CQ verification remains a significant barrier to adoption, creating an opportunity for more automated approaches that our work addresses.

With the advent of LLMs, several approaches have emerged to assist various ontology engineering tasks. LLMs have been employed to generate domain-specific CQs, as demonstrated by Di Nuzzo et al.~\cite{di2024automated}, who integrated GPT-4 with knowledge graphs to automatically generate domain-specific CQs and then used an LLM to judge their quality on criteria such as relevance, clarity, and depth. While promising, this work lacked comprehensive qualitative evaluation on a larger dataset (and comparison between LLM-generated metrics and human judgements) involving human engineers—a limitation our approach aims to overcome. In another work, Tüfek et al.~\cite{tufek2024validating} reported using GPT-4 to automatically convert CQs and other textual requirements into SPARQL queries for ontology validation. Similarly, Alharbi et al.~\cite{alharbi2024investigating} mentioned as a future direction of the work that LLMs could assist in determining whether a CQ truly aligns with an ontology's knowledge scope by translating CQs into SPARQL queries.

%In related work about ontology testing, other researchers have explored LLMs for verifying specific ontology constraints or definitions. For instance, Benson et al. \cite{} investigated how well GPT-4 could generate and evaluate ontology class definitions following an upper ontology standard (BFO). LLMs have begun to assist in evaluating ontology concept definitions for conformity to best practices. As Neuhaus (2023) observes, framing the bigger question of ontologies in the LLM era, the advent of LLMs opens the question of whether ontologies themselves should be adjusted in the era of LLMs, or vice versa, and how LLMs might validate ontological content.

Researchers have also begun exploring LLMs for ontology conceptualization evaluation. Tsaneva et al.~\cite{tsaneva2024llm} tested GPT-4 in the chat interface as an evaluator for ontology axioms compared to human experts. However, this kind of ontology evaluation is structural, not functional, and was carried out on one toy ontology only. Benson et al.~\cite{benson2024my} investigated GPT-4's ability to generate and evaluate ontology class definitions in the chat interface, following the Basic Formal Ontology. Despite suggesting that human-in-the-loop refinement shows potential for using GPTs as productivity tools for ontology-related tasks, the work was not tested on a larger-scale benchmark, but rather on selected example classes. Furthermore, the paper addresses structural correctness but not functional evaluation.

Thus, despite promising results, the use of LLMs for ontology evaluation, especially functional adequacy, remains underdeveloped. 
%As Neuhaus~\cite{neuhaus2023ontologies} frames the broader question of ontologies in the LLM era: the advent of LLMs opens inquiry into whether ontologies should be adjusted in the era of LLMs, or vice versa, and how LLMs might validate ontological content.
In this context, Garijo et al.~\cite{garijo2022llms} leave the categorisation of existing resources on LLM use for ontology evaluation blank, highlighting this gap.

Furthermore, current evaluation approaches for LLM-generated ontologies face significant methodological challenges. Lippolis et al.~\cite{lippolis2025ontology} proposed a comprehensive evaluation method, structural and functional, for LLM-generated ontologies that still, similar to Alharabi et al.~\cite{alharbi2024exploring}, relies heavily on manual components requiring the intervention of a domain expert. %The data produced in that work is the starting point for our OntAID dataset.
Similarly, Da et al.~\cite{da2024toward} developed LLM-based ontology generation techniques that depend on manual verification processes.

Rebboud et al.~\cite{RebboudLisenaTroncy2024} propose evaluation criteria including accuracy, completeness, and conciseness of generated ontologies measured against a ground truth. However, these criteria introduce two critical limitations: (i) they necessitate either a gold reference taxonomy or a comprehensive domain corpus—resources that are so far unavailable or incomplete, and (ii) they fail to explicitly incorporate ontology requirements in the evaluation process.

Also, a fundamental tension exists between expecting LLM evaluation to provide definitive answers versus using it as a decision support tool to assist human judgment. This distinction becomes crucial when establishing evaluation methodologies that balance automation with human-in-the-loop oversight.

Another important aspect in this field concerns the quality of requirements that seems to significantly impact ontology generation success  \cite{lippolis2025assessing}. Therefore, methods to evaluate ontologies against their intended functional requirements need to account for the quality of those requirements, especially when derived from multiple projects. To sum up, using LLMs as evaluators in ontology engineering is only now beginning to emerge. However, current LLM-based studies on ontology generation still rely on ad-hoc methods or measures requiring intense manual effort when performing a functional evaluation.

\section{The OntoEval dataset}
\label{sec:dataset}
This section describes the OntoEval dataset---annotated for ontology evaluation using CQ verification---by detailing its construction methodology and reporting its principal descriptive statistics. We then introduce OntoEval-small, a carefully curated subset of OntoEval designed for the semi-automatic ontology evaluation experiment with ontology engineers.

\subsection{Dataset creation}
\label{sec:dataset_creation}
%splitting it into two parts, Dataset Creation, dataset stats; there was no existing dataset, name it, this helps community: practical contribution. 
Our primary OntoEval dataset consists of 1,393 ontologies, both human-curated ones from European projects (57) and LLM-generated ontologies (1,336), each paired with a corresponding CQ, user story and annotations. Including potentially imperfect, LLM‑generated ontologies is intentional: real‑world evaluation tools must handle noisy or partially correct models, and such ontologies can provide controlled but realistic variation. OntoEval ontologies, stories and CQs are mostly derived from multiple recent works~\cite{lippolis2025assessing,lippolis2025ontology,saeedizade2024navigating}, in which ontologies were manually annotated for their evaluation with respect to the CQ Verification method into three labels: \textsc{Yes}, \textsc{No}, and \textsc{No minor}. The latter refers to the definition by \cite{lippolis2025ontology,saeedizade2024navigating} and indicates that only a single property is missing from the ontology to achieve complete CQ verification. In the automatic evaluation of this work, we treat \textsc{No minor} as \textsc{No}. Furthermore, CQs were assigned a difficulty rating on a binary scale, \textsc{Simple} or \textsc{Complex}, based on the presence of at most two classes and one object property for the \textsc{Simple} label, otherwise it was considered \textsc{Complex}. 
For instance, ``Who built and/or renovated an organ?'' is classified as \textsc{Simple} as it involves two classes (\texttt{Person}, \texttt{Organ}) and one object property (\texttt{built}/\texttt{renovated}): $\exists x, y : \text{Person}(x) \land \text{Organ}(y) \land (\text{built}(x,y) \lor \text{renovated}(x,y))$. Conversely, ``What was the disposition of the organ at a specific point in time?'' is \textsc{Complex} due to multiple classes and properties: $\exists x, y, z : \text{Organ}(x) \land \text{Parthood}(y) \land \text{TimeInterval}(z) \land \text{isWholeIncludedIn}(x,y) \land \text{hasTimeInterval}(y,z)$. 
Ontologies and their requirements were drawn from multiple sources to ensure a variety of domains and modelling patterns, and in particular, four large-scale European projects, one educational project with three ontologies, and three LLM-generated ontologies from recently peer-reviewed work:
\begin{enumerate}
    \item \emph{Polifonia Project} (domains: Music and Events)\cite{de2023polifonia};
    \item \emph{WHOW Project} (domains: Water and Health)~\cite{lippolis2025water};
    \item \emph{Onto-DESIDE Project}\footnote{\url{https://ontodeside.eu}} (domain: Circular Economy);
    \item IKS (Interactive Knowledge Stack for content management platforms)\footnote{\url{https://cordis.europa.eu/project/id/231527}};
    \item ``SemanticWebCourse'', used in Saeedizade and Blomqvist~\cite{saeedizade2024navigating} and Lippolis et al.~\cite{lippolis2025ontology}, consisting of three ontology stories about the music, theatre and hospital domain and 15 CQs each;
    \item AcquaDiva (domains: Microbe Habitat and Carbon and Nitrogen Cycling)\cite{algergawy2024towards};
    \item LLM generated ontologies in the works by~\cite{lippolis2025assessing,lippolis2025ontology,saeedizade2024navigating}.
\end{enumerate}

The LLM-generated ontologies were generated by the following LLMs: GPT-4 family (versions: GPT-4-0613, GPT-4-1106,
GPT-4-Turbo 2024-04-09, GPT-4-2024-05-13), Llama-3.1-405B-instruct, o1-preview, and DeepSeek-V3. In these works, a sample was evaluated, among other measures, by a panel of expert ontology engineers who found that while LLMs introduce distinctive modelling patterns, they do not necessarily underperform compared to novice human modellers. These ontologies were thus designed specifically to allow a controlled comparison between human-generated and LLM-generated artefacts using the same input requirements.

\subsection{OntoEval-small}

From the OntoEval corpus, we constructed a balanced subset for controlled experimentation, \textit{OntoEval-small}. To achieve this, we employed a combination of hard and soft constraints. As hard filters, we excluded ontology models addressing only minor issues (i.e., resulting in a binary \textsc{Yes}/\textsc{No} annotation) to make sure labels were unambiguous. Additionally, we retained only the CQs rated as \textsc{Complex} to have a more challenging evaluation. To ensure domain diversity, we limited the selection to a maximum of three CQs per project. LLM correctness served as a soft filter, meaning this filter was used as a sampling guide rather than a strict rule, such that the small dataset would have statistics similar to that of the full dataset with respect to correct and incorrect LLM predictions. We also sampled an approximately equal number of LLM-generated and human-curated ontologies and made sure that the LLM-generated ontologies were generated by different LLMs. Lastly, we selected a similar number of CQs that were modelled and not modelled. To facilitate user evaluation, each original story was condensed into a single-line summary using OpenAI's o1-preview model in a few-shot setting. %This process preserved core semantic content while improving readability.%the final dataset should contain similar statistics of LLM predicting them correct and incorrect in terms of automatic evaluation.

\subsection{Dataset statistics}
\label{sec:dataset_stats}
% Table~\ref{tab:ontoeval_combined} presents descriptive statistics for the two datasets used in our study: the full \textit{OntoEval} corpus and the smaller \textit{OntoEval-small} subset. The table represents statistics related to the total size of the dataset, the count of positive labels---correctly modelled CQs---the the remaining are negative, the number of CQs that are classified as simple based on the metric in \cite{saeedizade2024navigating,lippolis2025ontology}, domains that these ontologies belong to and the sources. It is worth mentioning that these negative classes are not generated by negative sampling techniques but extracted from future work of EU projects or manual. All LLM generated ontologies are manually annotated.

Table~\ref{tab:ontoeval_combined} summarises key descriptive statistics for \textit{OntoEval} and \textit{OntoEval-small}. For each dataset, we report the total number of CQs and the number of correctly modelled CQs. Most CQs in OntoEval are modelled and therefore the dataset is imbalanced (1,204 \textsc{yes} and 189 \textsc{no}). 
    %the count of positively labelled CQs (i.e., those correctly modelled in the ontology) and the remainder as negatively labelled CQs (i.e., not correctly modelled ones).
We also include the number of CQs classified as \textsc{Simple} and \textsc{Complex} according to the criteria outlined in Section~\ref{sec:dataset_creation},
    %in previous work~\cite{lippolis2025ontology,saeedizade2024navigating},
along with the domain coverage of the ontologies and their provenance. The ``not modelled CQs'' were not generated by automated negative‐sampling techniques but were instead drawn from planned EU project deliverables or manually annotated by ontology experts.
Furthermore, the ontologies are of different sizes, which are computed by the number of axioms in OntoMetrics~\cite{lantow2016ontometrics}, ranging from 15 to 567 axioms (mean: 132 axioms, median: 44 axioms, STD: 164 axioms).

    %experts during double-checking of the responses.
%All LLM-generated ontologies were additionally validated through manual verification. %In addition to OntoEval and OntoEval-small, we created another dataset completely separated from the datasets introduced, with another domain to select the final prompt and exclude some LLMs such as GPT-4.

\begin{table*}[t]
  \centering
  \caption{Comparison of the full \textit{OntoEval} corpus for automatic ontology evaluation and the \textit{OntoEval-small} subset for semi-automated evaluation.}
  \label{tab:ontoeval_combined}
  \begin{tabular}{lcc}
    \toprule
    \textbf{Metric} & \textbf{OntoEval} & \textbf{OntoEval-small} \\
    \midrule
    Total CQs                    & 1,393 & 20 \\
    % Unique CQs                   & 215  & 20 \\
    Modelled CQs                 & 1,204 & 10 \\
    % Non-modelled CQs             & 190  & 10 \\
    Difficulty: \textsc{Simple}             & 725  & 0  \\
    Difficulty: \textsc{Complex}             & 135  & 20 \\
    Domains         & 33   & 6  \\
    % LLM Eval: Right              & --   & 14 \\
    % LLM Eval: Wrong              & --   & 6  \\
    % Creator: GPT-4               & 742  & 1  \\
    % Creator: o1-preview          & 418  & 4  \\
    % Creator: Llama               & 90   & 3  \\
    % Creator: DeepSeek            & 88   & 2  \\
    % Creator: EU project               & 57 {\tiny(others LLMs)} & 10 \\
    % Source: Human curated & 57 & 10
  %& \makecell{57 \\ {\tiny others are LLMs generated}} 
  %& \makecell{10 \\ {\tiny}} 
    \bottomrule
  \end{tabular}
\end{table*}

\section{Methodology and experimental setup}
\label{sec:methodology}
\label{sec:experimentalsetup}

\begin{figure}
    \centering
    \includegraphics[width=1\linewidth]{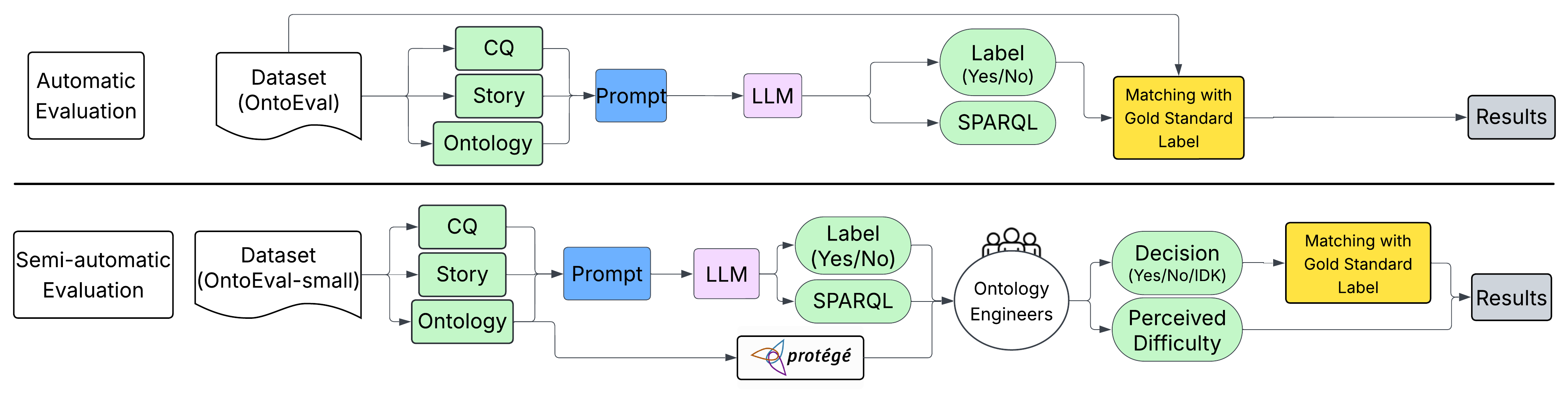}
    \caption{Overview of OE-Assist: (i) Automatic: predicted labels are compared to gold standards to compute performance metrics. (ii) Semi-automatic: Users aided by LLM suggestions assess whether a CQ is modelled.}
    \label{fig:map}
\end{figure}

In this section, we outline our dual‐phase OE-Assist methodology—comprising both fully automatic and semi‐automatic approaches—along with the corresponding experimental configurations, as shown in Figure \ref{fig:map}.

\subsection{Overall methodology}
OE-Assist---structured around semi-automatic and complementary automatic setups with ontology engineers---aims both to determine whether integrating AI-assisted suggestion can improve quality assessment and to explore the extent to which LLMs can independently evaluate ontologies as standalone agents. In this section, we describe the methodology of automatic ontology evaluation and the semi-automatic method.
%In the rest of this section, we describe the prompt and the experimental setup with respect to the automatic and semi-automatic evaluation of ontologies. 

%This dual-phase methodology provides a robust framework for assessing functional ontology quality: the automatic evaluation leverages LLM-generated labels for quick, scalable assessments, while the user-based evaluation introduces expert feedback to further refine and validate these outputs. %The integration of LLMs into ontology engineering environments, such as through Protégé, has the potential to not only expedite the evaluation process but also enhance the overall quality of ontological models and requirements.

\subsubsection{Automatic ontology evaluation.}
The automatic evaluation is designed as a label-based assessment mechanism. In this phase, a prompt is introduced to an LLM to analyse whether a CQ is adequately modelled within a given ontology and an ontology story. The evaluation is structured to classify ontology coverage without human intervention. The components of the prompt are the task description, CQs, a user story, and an ontology. The expected outputs are a binary class (\textsc{Yes}: CQ is modelled, \textsc{No}: CQ is not modelled) and a SPARQL query for CQ verification.

\subsubsection{Semi-automatic ontology evaluation.}

% The semi-automatic evaluation phase is focused on determining the extent to which LLMs can assist ontology engineers in improving the quality assessment process. Inputs and expected outputs of LLMs in semi-automatic are the same as the automatic method. Here, the label (Yes or No) and the SPARQL are presented to the user as suggestions. The user is expected to determine is a CQ is correctly modelled in the ontology or not, given an ontology, ontology story, and suggestions (label and SPARQL)

In the semi‐automatic evaluation, we investigate the extent to which LLMs can support ontology engineers in the ontology evaluation process. This phase employs the same inputs and expected outputs as the automatic approach. The LLM's suggested label (\textsc{Yes} or \textsc{No}) and the SPARQL query are provided to the user as suggestions. Given an ontology, its ontology story, and the LLM's suggestions, the user is responsible for determining whether each CQ has been accurately modelled within the ontology or not.

\subsection{Prompt design}

We propose a prompt\footnote{Due to space constraints, the full prompt specification is available on GitHub:\\ \url{https://github.com/dersuchendee/OE-Assist/tree/main/prompts}.} that (i) describes the ontology‐evaluation task, (ii) provides some examples (out of the OntoEval-small dataset, cf. Section ~\ref{sec:dataset_stats}) in a few-shot style, with both positive (\textsc{Yes}) and negative (\textsc{No}) cases, (iii) includes the user story, CQ, and ontology, and (iv) instructs the model to: (a) Return a Boolean label (\textsc{Yes} or \textsc{No}) indicating whether the CQ is correctly modelled, and (b) generate a SPARQL query to verify the CQ. %This order of outputs and the presence of examples were tested in another domain, not included in the provided and test dataset.
If the model returns \textsc{No}, it is also instructed to supply a partial SPARQL query highlighting the available portions of the ontology related to the CQ. %The prompt is formatted in a few‐shot style, incorporating illustrative examples of both positive (\textsc{Yes}) and negative (\textsc{No}) cases drawn from a distinct dataset mentioned in Section~\ref{sec:dataset_stats}. 

\subsection{Automatic ontology evaluation}

We automatically assess the correctness of each ontology using a set of three different LLMs on the OntoEval dataset and comparing the LLM \textsc{Yes}/\textsc{No} judgments against the gold standard.

%    \emph{[Prompt text to be inserted here.]}

\textbf{Models under test.}  
We run the prompt through three different LLMs: OpenAI o1-preview, GPT-4o-0513, and OpenAI o3-mini to represent a range of model capabilities and architectures. o1-preview was included as it is a state-of-the-art reasoning model with advanced capabilities for complex tasks. GPT-4o-0513 is a widely-used foundation model with established performance across knowledge-intensive tasks. o3-mini was included to test whether smaller, more efficient models could perform adequately on ontology evaluation tasks.
Each model is run three times, except o1-preview on the full OntoEval dataset and accessed via Azure API, with at least five hours between each run to avoid caching, using the default hyperparameters for o1-preview. Temperature and penalty are set to 0, along with keeping default hyperparameters, for others. %validation set is the dataset we used to find hyperparameters etc. Say there are few examples in it and in this section we say we used also gpt 4 1106 and deepseek etc and we say that we use models that perform better than the random baseline

\textbf{Answer extraction.}  
Although the prompt requests both the SPARQL query and the binary label, for this evaluation, we parse only the label to compute Macro-F1 and Accuracy by comparing it to the gold standard.

\subsection{Semi-automatic ontology evaluation}
This section presents the semi-automatic evaluation of ontologies. In this study, each participant was assigned 20 ontology evaluation tasks in a limited time. Half of these tasks included suggestions generated by an LLM, whereas the remaining tasks did not feature such assistance. The order of the tasks was changed between participants to mitigate the impact of learning effects on the results. The ontology stories, CQs, and the LLM-generated suggestions (if present) were shown using a basic user interface where ontology engineers could select a final assessment decision. A second screen provided access to the Protégé application with the ontology connected with the task pre-loaded. 
% \vspace{-1em}

\subsubsection{Participants.}
Nineteen ontology engineers participated in this evaluation. They were recruited through academic networks and represented roles ranging from research assistants to professors and affiliated with eight different international institutions with industrial or academic experience in semantic web technologies. Participants were divided into two experience groups (\emph{non-expert} vs.\ \emph{expert}) with professors---including assistant and associate professors---classified as \emph{experts}, and all others as \emph{non-experts}. While we acknowledge this distinction tends to simplify expertise, we opted for professors as a conservative and externally verifiable criterion over self-reported years of experience, which could yield inconsistent groupings. Seven participants completed the evaluation in an online setting, while the remaining twelve participated in person. %All participants had industrial or academic experience in ontology engineering and semantic web technologies.

\subsubsection{LLM selection.}
The LLM employed to generate suggestions (label and SPARQL) was o1-preview as it was the best performing one in the automatic ontology evaluation phase of the experiment.
% \vspace{-1em}

\subsubsection{Interface.}
A prototype of a web-based interface was developed to present LLM suggestions and requirements along with the ontology, as illustrated in Figure~\ref{fig:interface}. The interface was made available online for the duration of the experiment and was displayed side-by-side with Prot\'{e}g\'{e} to facilitate comparison. On the left, participants could browse the OWL ontology file in Protégé, while on the right they interacted with the evaluation interface. The interface displayed, for a given ontology, the current CQ and story, optional LLM suggestions, and response controls to answer if a CQ is modelled or not with the perceived difficulty on a Likert scale from 1 to 5. In this way, participants could carry out ontology evaluation in two modes: An assisted mode (with LLM suggestions) and an unassisted mode (without LLM suggestions).% Participants completed ontology evaluation tasks under both modes.% but on different tasks. 
% \vspace{-1em}

\subsubsection{Procedure.}
\begin{figure}[htbp]
  \centering
  \begin{subfigure}[b]{0.65\textwidth}
    \includegraphics[width=\textwidth]{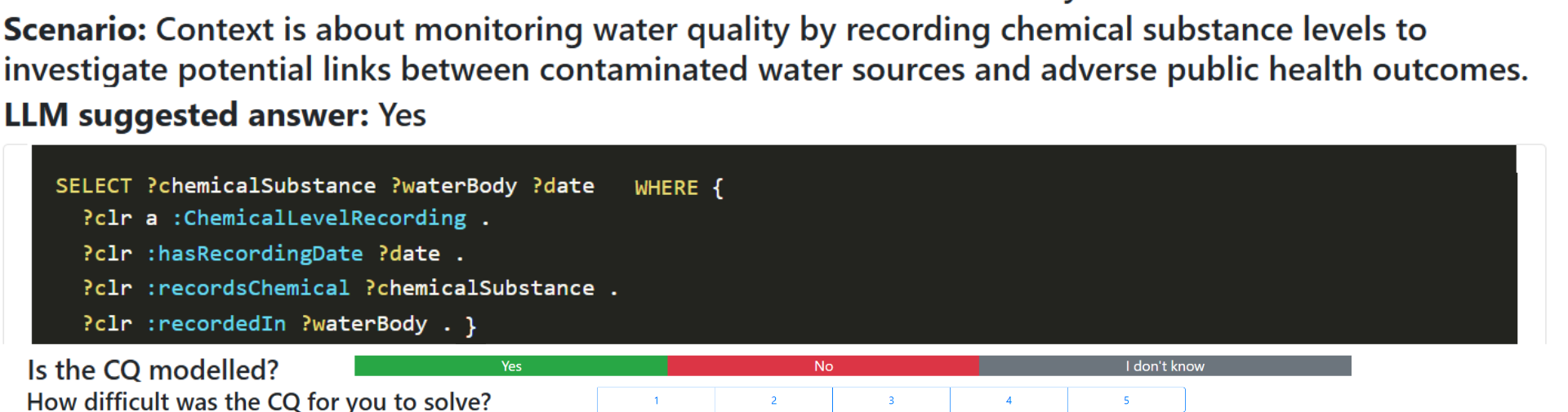}
    \caption{Assisted version in the interface.}
    \label{fig:interface1}
  \end{subfigure}
  \hfill
  \begin{subfigure}[b]{0.65\textwidth}
    \includegraphics[width=\textwidth]{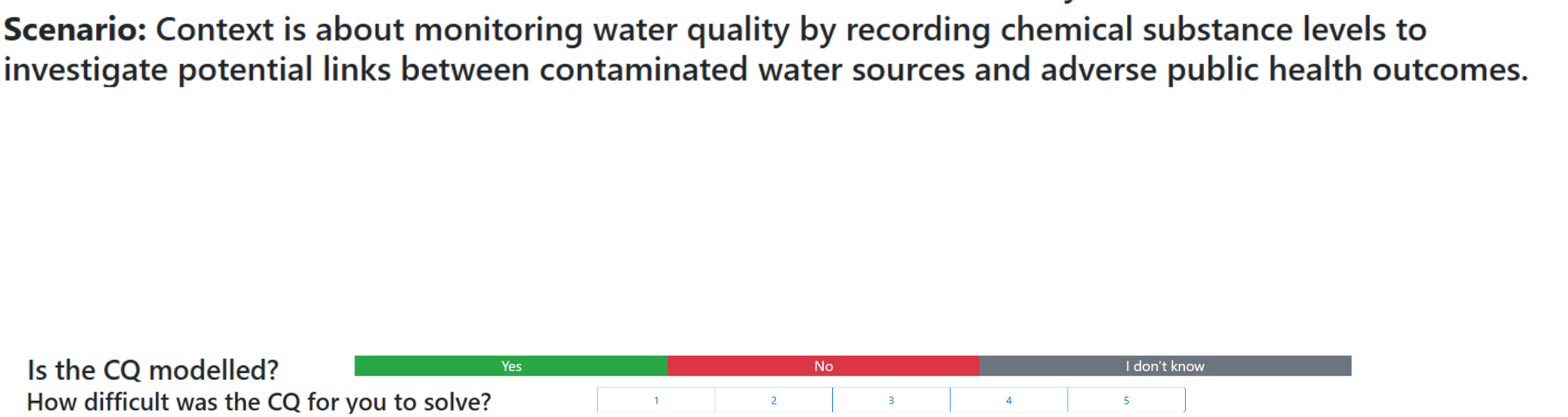}
    \caption{Unassisted version of the interface.}
    \label{fig:interface2}
  \end{subfigure}
  \caption{Comparison of assisted and unassisted settings in the interface.}
  \label{fig:interface}
\end{figure}

Participants completed two different sets of tasks (20 in total) under both conditions \emph{Assisted} (10) and \emph{Unassisted} (10) in a single session of approximately one hour. Prior to the main task, each participant completed a 10-minute brief tutorial, which included worked examples intended to familiarise them with the dual-window setup and the response workflow. Time was allotted for clarifying questions. Each condition utilised a dual-screen setup --- Prot\'{e}g\'{e} and the interface --- and was assigned a maximum time window of 20 minutes; CQs remaining at the end of this interval were recorded as skipped. Where participants were uncertain about a particular CQ, they could skip it by selecting ``I don’t know'' (IDK) via the interface. They could also skip each question if it was too difficult or if they could not understand a CQ. The expected sequence of actions was as follows: (i) participants were to open the ontology file in Prot\'{e}g\'{e} (without any plugins) and, after reading the competency question (CQ), the ontology story, and the suggestions (if available), (ii) they would select their answer (either ``Yes'', ``No'', or ``IDK'') and finally, (iii) indicate how difficult they found the task using a 1–5 Likert scale, as shown in Figure~\ref{fig:interface1} (assisted) and Figure~\ref{fig:interface2} (unassisted). Upon reaching the time limit or completing all tasks, participants switched to the alternate condition and proceeded with the remaining 10 questions. The sole distinction between conditions was the availability of LLM suggestions in the Assisted setup. A think‐aloud protocol was applied throughout the session, encouraging the participants to explain and comment on their actions and thoughts, to facilitate observational note-taking by the experiment observers. To conclude, participants completed a 10-item survey adapted from the System Usability Scale (SUS)~\cite{brooke1996sus}.%Specifically: %Only in the Assisted condition were LLM suggestions available.
%\begin{enumerate}
  %\item \textbf{Tutorial (up to 10 min)}: Participants received an overview of both setups, the CQ verification procedure, and the expected sequence of actions. Time was allotted for clarifying questions.
  
  %\item \textbf{Assisted and Unassisted Setups (2 × 20 min)}:  In each condition, participants attempted up to 10 CQs, indicating for each whether it was \emph{modelled}, \emph{not modelled}, or ``IDK'' followed by a perceived difficulty rating on a 1–5 Likert scale. Upon reaching the time limit or completing all tasks, participants switched to the alternate condition and proceeded with the remaining 10 questions. The sole distinction between conditions was the availability of LLM suggestions in the Assisted setup.

  %\item \textbf{Questionnaire (5 min)}: To conclude, participants completed a 10-item survey adapted from the System Usability Scale (SUS) \cite{brooke1996sus}.

%\end{enumerate}

%\subsubsection{Balancing the Tasks}
In our setup, we recognise that each ontology evaluation task could have different difficulty, so we made sure each CQ appeared equally in the assisted mode and unassisted mode. Also, the order of assisted mode and unassisted mode, whether it appears in the first half of the experiment or the second, is changed between participants to maintain a balanced distribution.

\vspace{-1em}

\section{Evaluation}
\label{sec:eval}

% \subsection{Evaluation}

In this section, we discuss evaluation metrics for both automatic and semi-automatic ontology evaluation.
% \vspace{-1em}
\subsubsection{Automatic ontology evaluation}
For OntoEval, we report macro‑averaged F1, which balances the majority (\textsc{Yes}) and minority (\textsc{No}) classes; for OntoEval-small, which is balanced, we report Accuracy and compare all LLMs with a uniform‑random baseline (Acc=0.50, F1=0.42).

%because the evaluation task is imbalanced. Macro‑F1 computes the F1 score independently for every class and then averages the scores, giving each class identical weight; this prevents the dominant \textsc{Yes} label from hiding poor performance on the rarer \textsc{No} label. Due to a lack of related work, all LLMs performances in ontology evaluation are compared to
%a random class prediction baseline
% two random baselines, one uniform and the other that predicts according to the empirical class prior distribution 
%(random with uniform: Accuracy = 0.50, F1 = 0.43).%; random choice with known class prior distribution: Accuracy = 0.758, F1 = 0.495).
% \vspace{-1em}
\subsubsection{Semi-automatic ontology evaluation}
\label{sec:semi-setup}

To assess the semi-automatic evaluation, we define each user’s accuracy in setup (assisted vs not assisted) \(m\) as \(\mathrm{Acc}_{u,m} = C_{u,m}/(C_{u,m}+I_{u,m})\), where \(C_{u,m}\) and \(I_{u,m}\) are the counts of correct and incorrect answers (``IDK'' ones are not counted here). We then compute the overall mode accuracy by averaging over \(N\) users: \(\mathrm{Acc}_{\mathrm{setup}} = \frac{1}{N}\sum_{u=1}^N \mathrm{Acc}_{u,\mathrm{setup}}\), and define \(\Delta\mathrm{Accuracy} = \mathrm{Acc}_{\mathrm{assisted}} - \mathrm{Acc}_{\mathrm{unassisted}}\). We also report the standard deviation of \(\mathrm{Acc}_{u,m}\) in each setup. To further analyse the relationship between LLM assistance and task correctness, we conduct a chi-square test of independence. Since each CQ was attempted multiple times by different users across both assisted and unassisted modes, we treat each answer as an independent observation and construct a 2×2 contingency table of assistance condition (present vs. absent) and answer correctness (correct vs. incorrect). %This allows us to test whether the availability of LLM suggestions significantly influences users' accuracy at the task level.

We used paired two-tailed t-tests to compare each user’s accuracy (and separately perceived difficulty) in the assisted vs. unassisted conditions, reporting the t-statistic and p-value. We assessed the relationship between perceived difficulty and accuracy using Pearson’s correlation coefficient, testing significance at $\alpha = 0.05$. Where relevant, we computed Cohen’s \textit{d} to quantify the magnitude of paired differences (e.g. in perceived difficulty or accuracy) \cite{becker2000effect}.

Finally, to investigate the potential impact of LLMs' correctness on user decision-making, we subdivide the reported results based on whether the LLM provides a correct suggestion or an incorrect suggestion. This approach allows us to discern how the varying degrees of LLMs' accuracy affect the overall effectiveness of the semi-automatic ontology evaluation methodology.

%For the semi-automatic evaluation of the ontologies, we report (i) statistical tests of data obtained in the assisted and non‑assisted settings and (ii) the metrics derived from the post‑session questionnaire, whose items were scored according to participants' responses. If a result is supported by the majority of the answered CQs—that is, by more than 10 CQs—we classify it as a finding. Support from 4 to 9 CQs is considered a tendency. If fewer than 4 CQs point to the same conclusion, the observation is ignored. Because the experiment was carried out under strict time constraints, participants were unable to address every CQ; this uneven coverage explains the prevalence of tendencies rather than confirmed findings in some cases. 

\section{Results}
\label{sec:results}
\vspace{-0.5em}
In this section, we present the results of the automatic and semi-automatic ontology evaluation.  We start with the automatic ontology evaluation on OntoEval and then select the best performing model to be used in the semi-automatic setting, assessing the users' performance in the various settings. 
% summary of what we are doing + say we discuss the findings
\vspace{-0.5em}
\subsection{Automatic ontology evaluation}
For the automatic evaluation of ontologies, as shown in Table~\ref{tab:eval-results}, o1‑preview overall outperforms the other LLMs on both datasets. 
The only weakness of o1‑preview was over human-curated ontologies, resulting in 0.05 lower score than o3-mini. Since all LLMs have a low standard deviation (STD) and the cost of running the experiment on o1-preview is high, we ran this model only once.
\vspace{-1.5em}
% ------------------------------------------------------------
% Automatic ontology‑evaluation results
% ------------------------------------------------------------
\begin{table}[h]
  \centering
  \caption{Performance of LLMs in the automatic ontology evaluation, compared to the random-guess baseline. Bold shows the best score on each column. %For semi-automatic ontology evaluation, o1-preview accuracy was 0.7.
           % $^\dagger$\;Accuracy is macro‑averaged over repeated CQs in the manually
           % double‑checked \emph{OntoEval‑small} subset.
           }
\resizebox{.85\textwidth}{!}{
\begin{tabular}{llll l}
  
    \toprule
    & \multicolumn{3}{l}{Macro‑F$_1$ (on OntoEval)} & Accuracy (on OntoEval‑small) \\[-1pt]
    \cmidrule(lr){2-4}\cmidrule(l){5-5}
    \textbf{Model} & EU projects, & LLM Generated, & All & All%OntoEval‑small%$^\dagger$ 
    \\  
    \midrule
    Random baseline & 0.41 & 0.42 & 0.43 & 0.50 \\
    
    GPT‑4o‑0513     & 0.34 $\pm 0.02$ & 0.48 $\pm 0$& 0.48 $\pm 0$& 0.55 $\pm 0$\\
    
    o3‑mini         & \textbf{0.51} $\pm 0.02$& 0.56 $\pm 0.01$& 0.58 $\pm 0.01$& 0.72 $\pm 0.02$ \\
    
    o1‑preview      & 0.46 & \textbf{0.66} & \textbf{0.66} & \textbf{0.75} $\pm 0.05$ \\
    \bottomrule
  \end{tabular}}
  
           \label{tab:eval-results}
\end{table}

\vspace{-1.5em}

\subsection{Semi-automatic ontology evaluation}

In this section, we present the results for the semi-automatic ontology evaluation, focusing on the differences in accuracy and perceived difficulty in the assisted compared to the unassisted mode. We also examine how the presence of suggestions from LLMs—whether correct or incorrect—shape these outcomes. In addition, we evaluate the relationship between perceived difficulty and correctness. Finally, we discuss the observed learning curve and the questionnaire that users filled in, highlighting how their responses inform the broader evaluation of ontologies and the role of LLMs within this process.

\vspace{-1.5em}
\subsubsection{Accuracy in the assisted compared to the unassisted setting.}
Using the metrics described in Section~\ref{sec:semi-setup}, we calculated accuracy in both the assisted and unassisted settings. We then computed the difference ($\Delta\text{Accuracy}$) for each condition, taking into account whether the suggestion provided was correct or not. By analysing the relationship between these variables, we assessed whether the presence of LLM suggestions had a statistically significant effect on correctness.

\begin{figure}[htbp]
  \centering
  \begin{subfigure}[b]{0.3\textwidth}
    \centering
    \includegraphics[width=\textwidth]{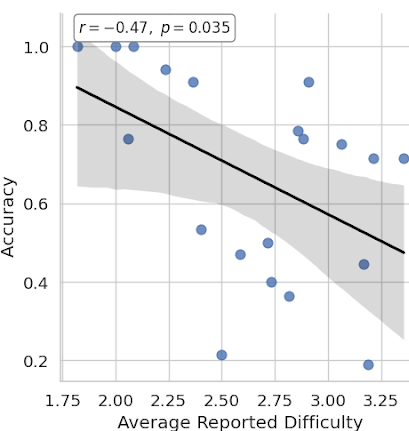}
    \caption{Difficulty vs Accuracy.}
    \label{fig:ontoevaal-perceiveddiff}
  \end{subfigure}\hfill
  \begin{subfigure}[b]{0.7\textwidth}
    \centering
    \includegraphics[width=\textwidth]{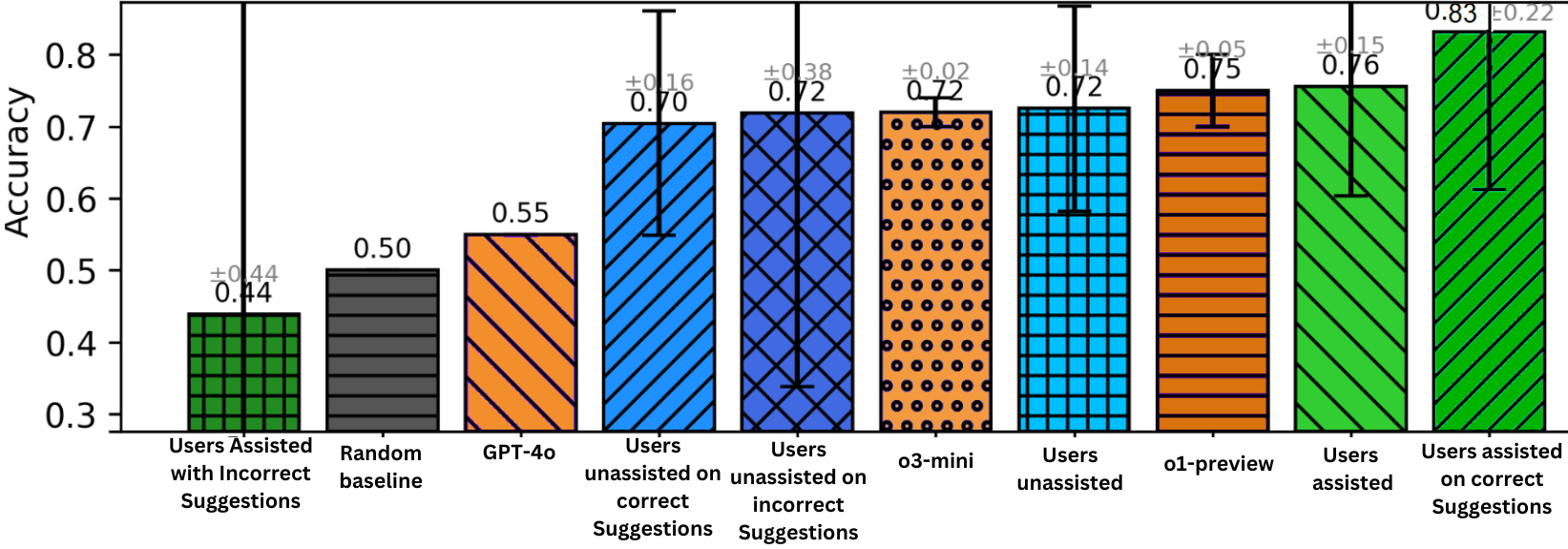}
    \caption{Accuracy of LLMs and users on different parts.}
    \label{fig:acc}
  \end{subfigure}
  \caption{Results overview. a) Relationship between difficulty and accuracy: easier perceived tasks are solved more accurately; (b) Comparative accuracy of the LLMs (orange) and human evaluators (green: assisted and blue: unassisted), with user responses further stratified by LLM correctness. STD is in grey.}
  %\label{fig:interface}
\end{figure}

As accuracy scores are illustrated in Figure~\ref{fig:acc}, when the LLM provided correct suggestions, we observed a significant average improvement of +13\% (from 70.46\% to 83.18\%), supported by a paired t-test (t = 2.04, p = 0.047). However, when the LLM suggestions were incorrect in the same task, performance significantly declined with an average change of -28\% (from 71.93\% to 43.86\%), confirmed by a t-test (t = -2.09, p = 0.043).
Since correct LLM suggestions occurred more frequently overall, there was a total improvement of +0.04\% (0.72 to 0.76) for LLM-assisted users (see Figure~\ref{fig:acc}).
However, this difference was not statistically significant. In addition, o3-mini and o1-preview have a similar accuracy to users' performance in both modes.
%However, the t-test gives a high p-value,
%meaning we cannot reject the claim that there is no difference between groups. Therefore, the improvement could not be conclusively confirmed or rejected.
% However, the t-test failed due to a high p-value, indicating that the hypothesis of a net performance shift, aggregated over all tasks could not be conclusively confirmed or rejected.

Using a chi-squared test, we found statistically significant associations for both correct ($\chi^2$ = 4.12, p = 0.042) and incorrect ($\chi^2$= 4.61, p = 0.0318) LLM suggestions. These findings underscore that the presence of LLM input, within the experimental setup, for each CQ, indeed affects decision outcomes. %Furthermore, there is no statistical difference between experts and non experts' performance.

\vspace{-1.5em}

\subsubsection{Perceived difficulty in assisted compared to unassisted mode.}

Participants judged the tasks as slightly less difficult when LLM suggestions were present (mean difficulty = 2.52, STD = 0.56) compared to when they were absent (mean difficulty = 2.75, STD = 0.46), where mean difficulty represents the average difficulty rating across all CQs on a Likert scale, and STD indicates how much those ratings varied from one CQ to another. This difference is small (Cohen’s \textit{d} $\approx$ 0.22), indicating a modest effect size; in behavioral research, \textit{d} values around 0.2 typically reflect small but meaningful differences. Note that this reflects variation across CQ-level averages rather than individual tasks.

As expected, experts consistently rated tasks as easier than novices, with the average difficulty dropping by 9\% for experts (from 2.67 to 2.44) and 5\% for novices (from 2.84 to 2.69) when assisted. Thus, both groups experienced slightly lower perceived difficulty with suggestions, but the expert group consistently found the tasks easier overall. For what concerns ``I don't know'' responses, users settled on ``I don't know'' almost exclusively when they perceived the question as hard (mean difficulty=3.9 vs 2.4 for other responses). Furthermore, if the accuracy of the CQ was low, there was also a higher number of ``I don't know'' responses. Instead, ontology size in terms of number of axioms does not play a role in pushing people towards being undecided (r=0.01, p=0.8).%: more likely, their decision is driven by their subjective difficulty towards, for instance, unclear phrasing or domain-specific CQs.
\vspace{-1.5em}
\subsubsection{Relationship between perceived difficulty and accuracy.}
Figure \ref{fig:ontoevaal-perceiveddiff} depicts Pearson's correlation between perceived difficulty and accuracy in our setup. There is a moderate negative correlation (r = -0.47, p = 0.035) between perceived difficulty and accuracy, indicating that as users perceived tasks to be harder, their accuracy tended to decrease. This correlation seems to be explained by the tasks in which the LLM gave incorrect suggestions—users found those tasks harder and were also less likely to solve them correctly. Furthermore, when the tasks were analysed separately by the correctness of the LLMs’ suggestions, the difficulty–accuracy link was no longer significant (LLM correct: r = -0.23, p = 0.44; LLM incorrect: r = +0.21, p = 0.69).

% \begin{figure}[h]
%     \centering
%     \includegraphics[width=0.4\textwidth]{Images/ontoevaal-perceiveddiff.png}
%     \caption{Perceived difficulty vs. accuracy for each CQ. Points show average difficulty (x-axis) and accuracy (y-axis) across participants.}
%     \label{fig:ontoevaal-perceiveddiff}
% \end{figure}
\vspace{-1.5em}

\subsubsection{Learning effect curve by expertise.}

As participants progressed, they rated each successive task as easier, indicating a learning effect. Participants who began in the assisted mode showed almost no change when they later switched to the unassisted mode (-0.04 accuracy), whereas those who began unassisted improved substantially after switching to assistance (+0.17 accuracy). Expertise further modulated this pattern: non-experts gained little when assisted first, yet improved modestly without assistance (+0.06). In contrast, experts improved sharply in the unassisted block (+0.29) and only slightly with assistance (+0.06). Across all users, accuracy rose in the second half for 40\% of CQs solved without LLM support but for just 15\% with support. A paired‑samples test approached significance (p$\approx$0.08), and the effect size (Cohen’s \textit{d}$\approx$0.45) indicates a medium learning advantage in the unassisted condition—but definitive statistical confirmation requires a larger sample.

%We observe a difference in learning trends between assisted and unassisted conditions. Since each user started with different mode ---assisted or unasssited--- they were performing differently on the second half of the experiment. On average, users who started with assisted mode performed similarly in the unassisted mode ($0.04$ less accurate in the second half) but when they started with unassisted mode, $+0.17$ improvement was seen. This shows with LLM suggestions users' need for learning the task is reduced.
% On average, accuracy decreased slightly in the assisted setting ($-0.04$), while it increased in the unassisted one ($+0.17$). 
%When broken down by expertise, novices showed a small improvement without assistance ($+0.06$) but no gain when assisted first. Experts, in contrast, showed a marked improvement without assistance ($+0.29$) and a smaller gain with assistance ($+0.06$). This suggests a learning effect primarily in the unassisted condition. Specifically, accuracy improved in the second half of the session for 40\% of the CQs with no LLM assistance, compared to only 15\% of CQs in the assisted condition. Despite the paired‑sample test's result where p‑value is \~.08, its Cohen’s \textit{d} of about 0.45 indicates a medium practical effect—the learning advantage without assistance is sizeable but the sample is just a bit too small to reach conventional statistical certainty.
%: the presence of LLM support may have reduced opportunities for users' improvement over carrying out the task
\vspace{-1.5em}
\subsubsection{Questionnaire.}

In the post‑task questionnaire, 63\% of the participants strongly agreed that the LLM suggestions were useful in solving the task, 74\% felt they could learn to use the system setup quickly,
    %79\% stated either the suggestions, the suggested label or the SPARQL queries, were not distracting.
and, 21\% found either the suggestions, the suggested label, or the SPARQL queries distracting.
%This suggests users would likely want to have access to a tool to get support for this type of ontology evaluation task.
%Opinions were more divided on the final yes/no recommendation feature: only 32\% endorsed it, while 47\% disagreed, pointing to a need for a clearer decision-support rationale.

\section{Discussion}
\label{sec:discussion}
This section discusses our key findings and situates them within the broader context of automated and semi-automated ontology evaluation.

\textbf{Automatic Ontology Evaluation.}
From our results, GPT‑4o hover around the random baseline, whereas o1‑preview delivers~+0.23 F1 over random on automatically generated ontologies and~+0.15 on human curated ones. Similar improvement was seen for o3-mini, suggesting that these reasoning LLMs can be helpful for ontology evaluation, especially given the steady upward trend in LLM performance. These results imply that o1‑preview and o3‑mini can already serve as an automated first-pass layer flagging potential errors before human review; for instance, they could accelerate curation in large, rapidly updated ontologies. %Fine‑tuning such models on domain‑specific CQs could push accuracy further, making them an even more reliable component of continuous integration pipelines. %Incorporating o1-preview into a fully automated ontology‑generation pipeline should both raise precision and reduce curation cost. 

% \subsection{Semi-automatic Ontology Evaluation}

% In this section, we discuss the semi-automatic evaluation of ontologies with respect to general benefits and drawbacks, ontology engineers' strategies for assisted settings, limitations and future work.

% \subsubsection{Benefits and drawbacks of OE-Assist}
\textbf{Semi-automatic ontology evaluation.} Our findings show that while LLM-assisted ontology evaluation improves the accuracy by 13\% when the suggestions are correct, the approach exhibits a significant decline when the user is supplied with incorrect suggestions. Overall however, these opposing effects largely cancelled each other out, resulting in no statistically significant net change in overall accuracy. Given that the o1-preview model’s suggestions used in the experiment had a  70\% accuracy, we speculate that future LLMs—with higher suggestion accuracy rates—will enable end
users to realize a correspondingly greater enhancement in task performance. One may also consider other ways to mitigate the effects of wrong LLM suggestions, such as pre-checking suggested SPARQL queries against the ontology before showing the suggestion to the user, or including some kind of confidence assessment of the suggestions.

Also, LLM suggestions reduced the perceived difficulty (mean = 2.52 vs.\ 2.75). One downside of the assisted condition, however, is that it may hinder learning over time. The modest learning gains in unassisted settings suggest that task complexity from the user's point of view is higher without automated support (accuracy improved of 40\% for CQs without LLM suggestions in the second half of the experiment vs.\ 15\% with suggestions). This learning effect was shaped by task order: users who started with assistance received early support and thus performed equally well in the later unassisted phase as those who learned through doing. As a result, overall learning gains appear neutralised—since early assistance seems to substitute the effort that would otherwise lead to similar improvement.

Ontologies contained in OntoEval-small had different sizes. However, we found there is no correlation between the size of ontologies in terms of axiom number and incorrect answers.
However, we note that even the largest ontology used is small by real-world standards. Taken together, these results imply that LLM support seems to be valuable mainly when its precision is high and that future versions of the prototype should pair assistance with dedicated safeguards—and perhaps dedicated learning phases—to avoid over‑reliance on LLMs and preserve users’ skill growth, but also that in realistic scenarios, automated suggestions could become even more valuable.
%Furthermore, there is no correlation between larger ontologies in terms of axiom number and perceived difficulty (r=0.04, p=0.54).

% The divided stance on recommendation functionality points to a need for strengthening the decision-support rationale, which however was not included in the design of the experiment as it would have increased the time and complexity of the setting.

\textbf{Feedback during the experiment.} % During and after the experiment, thoughts out loud of users could help us understand more about their usage of OE-Assist for what concerns both strategies of use and quality of requirements. % \paragraph{Strategies on the assisted setting.}
Looking at users' behaviour performing the tasks revealed %two distinct user approaches 
that when working with the assisted setting many users started the evaluation from the SPARQL queries, to find and match the named classes and properties with the ones mentioned in the CQ. A second group of users operated in reverse: first exploring the ontology through Protégé to establish a conceptual understanding, and only after that they proceeded to study the suggested SPARQL query.
These contrasting workflows suggest different methods of problem-solving—one favoring a code-first approach that uses the ontology as confirmation, and another prioritizing conceptual understanding before proceeding with CQ verification. 
In line with these strategies, in the questionnaire feedback, users who took the code‑first path appear to have leaned heavily on the system’s inline suggestions to speed-check query fragments—an affinity that helps explain why 63\% strongly agreed the suggestions were useful. 
Where the two groups converged was on the final decision aid. Only 32\% endorsed the binary \textsc{Yes}/\textsc{No} recommendation. Indeed, user feedback revealed challenges with the lack of labels and comments in some ontologies, unclear CQs, domain-specific content, and flawed LLM-generated hierarchies, all of which required extra effort to interpret and evaluate. These findings highlight the need for adaptable AI interfaces that support both code-first and concept-first user strategies, while emphasizing the importance of providing transparent, context-rich justifications in addition to binary recommendations. 

\textbf{Quality of requirements.}
 User comments and the varying wording of the CQs show that inconsistent, under‑specified, or overly domain‑specific requirements impede reliable evaluation. When the wording or granularity of the questions shifted from one task to the next, participants reported uncertainty about how to assess the related ontology. Domain specificity created an additional hurdle: evaluators who lacked subject‑matter expertise could judge structural soundness but not semantic correctness, showing LLM suggestions seem to not be helpful in these cases. %In short, clear and consistently scoped requirements are a prerequisite for trustworthy quality assessments—especially when evaluators and knowledge engineers do not share the same domain background.

% mix it with the questionnaire

%\subsubsection{Challenges.}
%During the evaluation, some challenges have emerged via the users' comments. The lack of descriptive labels in some ontologies forced participants to rely on their intepretations of some domain-specific concepts. Domain specificity presented another obstacle, as users without subject matter expertise struggled to assess the semantic correctness of generated ontologies despite understanding their structural aspects. Additionally, LLM-generated ontologies occasionally exhibited problematic taxonomic hierarchies with inappropriate subclass relationships, redundant classifications, or misplaced properties, requiring users to invest additional effort in reorganizing these conceptual structures for proper domain representation.
%\textbf{Impact of ontology size.}
%In the experiment, ontologies had different sizes. However, we found there is no correlation between larger ontologies in terms of axiom number and incorrect answers (r=0.01, p=0.8).
%Furthermore, there is no correlation between larger ontologies in terms of axiom number and perceived difficulty (r=0.04, p=0.54).

\textbf{Limitations and risks.} While our findings indicate that LLMs hold considerable promise for user-based functional ontology evaluation, future work should involve a broader group of ontology engineers and a wider range of LLMs to validate these results, since our group included only five participants with expert qualification level. Additionally, some participants had worked on the ontologies several years earlier, so even though they reported remembering little, residual familiarity could still have influenced their answers. Finally, our study highlights the need to improve the quality of the requirements and to integrate the workflow into widely used tools, such as Protégé, in order to enhance overall usability.
For dataset leakage, we did not find ontologies and their CQ annotations co‑located online, so direct memorisation by the LLM is unlikely. Still, because pre‑training data are opaque, we cannot rule out partial leakage.
Running a full automated evaluation with o1-preview cost approximately \$851. By contrast, a single run with gpt-4o cost about \$40, and o3-mini just \$25, so we only ran o1-preview once. These differences highlight a trade-off between model strength and budget that must be considered when selecting engines or planning multiple runs.

\vspace{-1.3em}
\section{Conclusion}
\label{sec:conclusion}
In this work, we have introduced
%We introduced OE‑Assist,
the first end‑to‑end framework using LLMs for functional ontology evaluation via CQ verification through both automatic and semi-automatic methods. We also introduced OntoEval, a benchmark of 1,393 ontology–CQ pairs across 33 domains, annotated by ontology engineers. In response to RQ1, ``To what extent can LLMs evaluate ontologies using CQ verification?'', in automatic experiments, the o1‑preview model achieved 0.66 macro‑F1, outperforming other LLMs. Hence, we conclude that newer LLMs perform on par with human users. To evaluate practical impact of the semi-automated methodology, and explore RQ2 (``To what extent can LLMs assist ontology engineers in evaluating
ontologies through CQ verification, and what are the benefits and drawbacks
of a hybrid approach combining LLM suggestions with expert validation
compared to traditional human-only methods?''), we ran a controlled study with 19 ontology engineers from eight institutions. When o1‑preview's suggestions were correct, average user accuracy improved by 13\% and perceived task difficulty decreased. However, incorrect suggestions caused a 28\% drop in accuracy; resulting in the same overall accuracy as unassisted mode since correct suggestions outnumbered incorrect ones. We conclude that the usefulness of LLM assistance heavily depends on the suggestion accuracy, and users are easily influenced by erroneous suggestions, while still perceiving the overall approach as useful. As LLMs improve, these results bode well for reasoning-powered ontology evaluation tools.%As newer LLMs further improve performance, these results indicate strong potential for LLM‑powered ontology evaluation tools with reasoning LLMs.

%Future work will expand OE‑Assist with richer, evidence-driven suggestions, enabling LLMs to extract relevant ontology fragments for visualisation and to provide explanations of their decisions. It will also investigate fine‑tuning and evaluation on larger multi-module ontologies.

\textbf{\textit{Reproducibility statement:}} Code, data, and supplementary material used in this paper are available on GitHub: \url{https://github.com/dersuchendee/OE-Assist}.

\begin{credits}
\subsubsection{\ackname}
\footnotesize
This project has received funding from the European Union’s Horizon Europe research and innovation programme under grant agreements no. 101058682 (Onto-DESIDE) and 101070588 (HACID), and is supported by the strategic research area Security Link. The data from the ``SemanticWebCourse'' used in the research were collected as part of a master's course taught by Assoc. Prof. Blomqvist while employed at Jönköping University.
Additional financial support to this project was provided by NextGenerationEU under NRRP Grant agreement n. MUR IR0000008 - FOSSR (CUP B83C22003950001).
This work was also supported by the PhD scholarship ``Discovery, Formalisation and Re-use of Knowledge Patterns and Graphs for the Science of Science'', funded by CNR-ISTC through the WHOW project (EU CEF programme - grant agreement no. INEA/CEF/ICT/ A2019/2063229). We thank OpenAI's Researcher Access Program Grant for the API credits. Finally, and most importantly, we deeply thank all participants that helped evaluation in this work's experiments.
\end{credits}
\normalsize

%\begin{credits}
%\subsubsection{\ackname} A bold run-in heading in small font size at the end of the paper is
%used for general acknowledgments, for example: This study was funded
%by X (grant number Y). We also thanks evaluators....to be done if accepted

%\subsubsection{\discintname}
%Thanks to evaluators, projects and?
%\end{credits}
\bibliographystyle{splncs04}
\bibliography{bibliography}

\end{document}